\documentclass{article}
\usepackage[margin=1in]{geometry}

\usepackage{amsmath}
\usepackage{graphicx}

\usepackage[labelsep=endash, labelfont=bf, format=plain, textformat=simple, justification=justified, textfont=small]{caption}
\usepackage[labelfont={bf, large}, position=top, singlelinecheck=off, justification=justified, labelformat=simple, skip=0pt]{subcaption}
\DeclareCaptionLabelFormat{bold}{\textbf{#2}}
\usepackage[utf8]{inputenc} 
\usepackage[T1]{fontenc}    
\usepackage{hyperref}       
\usepackage{url}            
\usepackage{booktabs}       
\usepackage{amsfonts}       
\usepackage{nicefrac}       
\usepackage{microtype}      

\usepackage{algorithmic}
\usepackage[ruled,noend]{algorithm2e}
\newcommand{\RR}{\mathbb{R}} 
\newcommand{\Fcal}{\mathcal{F}}
\newcommand{\Ical}{\mathcal{I}}
\newcommand{\Mcal}{\mathcal{M}}
\newcommand{\Wcal}{\mathcal{W}}
\DeclareMathOperator{\Tr}{Tr}

\usepackage{wrapfig}
\def\abovestrut#1{\rule[0in]{0in}{#1}\ignorespaces}

\date{\today}

\begin{document}

\title{Supervised Quantile Normalisation}
\author{Marine Le Morvan$^{1,2,3}$ and Jean-Philippe Vert$^{1,2,3,4}$\\ \\
$^1$MINES ParisTech, PSL Research University, CBIO-Centre for Computational Biology,\\
75006 Paris, France\\
$^2$Institut Curie, 75248 Paris Cedex 5, France\\
$^3$INSERM, U900, 75248 Paris Cedex 5, France\\
$^4$Ecole Normale Sup\'erieure, Department of Mathematics and Applications, Paris, France\\ \\
\url{{marine.le_morvan, jean-philippe.vert}@mines-paristech.fr}
}

\maketitle

\begin{abstract}
Quantile normalisation is a popular normalisation method for data subject to unwanted variations such as images, speech, or genomic data. It applies a monotonic transformation to the feature values of each sample to ensure that after normalisation, they follow the same target distribution for each sample. Choosing a "good" target distribution remains however largely empirical and heuristic, and is usually done independently of the subsequent analysis of normalised data. We propose instead to couple the quantile normalisation step with the subsequent analysis, and to optimise the target distribution jointly with the other parameters in the analysis. We illustrate this principle on the problem of estimating a linear model over normalised data, and show that it leads to a particular low-rank matrix regression problem that can be solved efficiently. We illustrate the potential of our method, which we term SUQUAN, on simulated data, images and genomic data, where it outperforms standard quantile normalisation.
\end{abstract}

\section{Introduction}

In many application fields where data are collected for a particular task, data acquisition is often plagued with various sources of perturbations which induce unwanted variations in the captured data and make the task harder to solve. For example, two photos of the same object taken from the same position may still vary considerably in terms of color distribution or other statistical properties depending on the ambiant light, the device used to take the picture, or the person in charge of taking the picture \cite{Gonzalez2008Digital}. Similarly, pixel intensities of an MRI scan do not have a fixed meaning and can vary considerably between two scans on the same patient with the same protocol and same scanner \cite{Shinohara2014Statistical}; speech recognition is challenging in part because the acoustic signal corresponding to a given word varies a lot with the speaker, the noise pollution around and the device used to capture the signal \cite{Hilger2006Quantile}; and microarray - or sequencing - based measurements in genomics are famous for being extremely sensitive to a variety of unwanted perturbations such as temperature, sample preparation protocol, or amount of material \cite{Bullard2010Evaluation}.

In order to reduce the burden of unwanted variations for subsequent data analysis applications, the standard way to proceed is often to \emph{normalise} the data prior to any analysis, in order to remove unwanted variations as much as possible while keeping relevant signals. Normalisation procedures vary from simply centering and scaling each sample to impose a common scale across samples, to more sophisticated and data-specific procedure, e.g., \cite{Bullard2010Evaluation}. In this work we are interested in a particular normalisation procedure, pervasive across different fields and known under different names, which monotonically modifies the entries of a given sample so that after normalisation, all samples have the same distribution of entries. Following the terminology used in biostatistics \cite{Hicks2015quantro}, we refer to this procedure as \emph{quantile normalisation} (QN). QN is ubiquitous in high-dimensional biological data analysis, where samples are often corrupted by various technical or biological unwanted variations, and is widely used for many types of data including low-density \cite{Amaratunga2001Analysis,Yang2003Statistics} or high-density \cite{Bolstad2003comparison,Irizarry2003Exploration} microarray data for gene expression analysis, high-density microarray for genotyping \cite{Carvalho2007Exploration,Scharpf2011Using}, RNA-seq sequencing data for gene expression analysis \cite{Cloonan2008Stem,Bullard2010Evaluation,Dillies2013comprehensive}, microarray data for DNA methylation analysis \cite{Yousefi2013Considerations}, or ChIp-seq sequencing data for protein-DNA interaction analysis \cite{Bilodeau2009SetDB1,Kasowski2010Variation}. QN is also widely used in image processing under the name of \emph{histogram matching}, or more specifically \emph{histogram equalisation} when the pixel intensities of an image are monotonically transformed in such a way that the distribution of values becomes approximatively uniform \cite{Gonzalez2008Digital}. A popular application of histogram matching is in MRI brain imaging, where a popular approach to preprocess images is to apply a variant of QN proposed by \cite{Nyul1999standardizing} and refined by \cite{Nyul2000New} and \cite{Shah2011Evaluating}. Similarly, another variant of QN targeting a uniform distribution is popular in speech recognition under the name of \emph{histogram normalisation} \cite{Dharanipragada2000nonlinear,Molau2001Histogram,Hilger2006Quantile}. In geostatistics, a popular trick to analyse non-gaussian spatial data is to perform a \emph{Gaussian anamorphosis}, i.e., a QN where the data is modified to follow an approximately gaussian distribution \cite{Chiles2012Geostatistics}.

In spite of its popularity and success, QN suffers from a practical question: \emph{how to choose the target distribution?} Various choices of target distribution have been popularised for different reasons in different fields, such as the uniform distribution in histogram equalisation in order to increase the global contrast of images; the gaussian distribution in Gaussian anamorphosis in order to be able to apply statistical methods that work well for gaussian data; or the median of the empirical distribution of the samples in biology as an attempt to keep some information of the original values. Beyond such heuristics, we are not aware of any rigorous guiding principle that could justify these choices, and as mentioned by Bolstad et al. \cite{Bolstad2003comparison}, "it seems unlikely that an agreed standard could be reached" for the choice of the target distribution, leaving this question largely open.

In this work we propose a general principle to answer this question, namely, \emph{to optimise the target distribution for the task to be performed after normalisation}, and illustrate this principle when after normalisation a linear model is trained for a classification or regression task. Coupling prior normalisation with subsequent linear model estimation results in a new model, which we term \emph{supervised quantile normalisation} (SUQUAN), where the optimal target distribution is the solution to an optimisation problem. We show that, equivalently, SUQUAN can be thought of as a particular linear model with rank constraint over the space of $p\times p$ matrices, where each sample $x\in\RR^p$ is embedded as a permutation matrix defined by the order of its features. We propose three algorithms to approximate a solution under different prior assumptions on the target distribution. We illustrate the behavior of SUQUAN on simulated data and on real images and biological data, where it outperforms the standard QN procedures.

\section{Quantile normalisation (QN)}

Let us first set up notations and present the standard QN procedure. We consider data $x_1, \ldots, x_n \in\RR^p$ where each sample is a $p$-dimensional vector, such as an image represented by the intensities of $p$ pixels or a biological sample represented by the expression of $p$ genes. QN is a nonlinear transform $\Phi_f : \RR^p \rightarrow \RR^p$ indexed by a vector $f\in\RR^p$ which we call the \emph{target quantile}. In words, QN monotonically modifies the entries of any input vector $x$ so that $\Phi_f(x)$ has the same distribution of entries as $f$, but ranked in the same order as the entries of $x$. When $f=(f_1,\ldots,f_p)^\top$ is a valid quantile its entries  are sorted in increasing order $(f_1\leq f_2 \leq \ldots \leq f_p$), so that the smallest entry of $x$ becomes $f_1$ in $\Phi_f(x)$, the second smallest becomes $f_2$, and so on. Ties in the entries of $x$ are arbitrarily broken, e.g., by considering $x_i$ before $x_j$ if $x_i=x_j$ and $i<j$.

QN can be formalised mathematically as follows. Given any $x\in\RR^p$, we call $\Pi_x$ the $p\times p$ binary permutation matrix defined by $(\Pi_x)_{ij}=1$ if the $i$-th entry of $x$ is ranked at the $j$-th position when all entries of $x$ are sorted from the smallest to the largest. Then by construction, the QN normalisation can be simply written as:
\begin{equation}\label{eq:qn}
\forall x\in\RR^p\,,\quad \Phi_f(x) = \Pi_x f \,.
\end{equation}
The following example illustrates these notations and the relation (\ref{eq:qn}) for an arbitrary sample $x\in\RR^4$ and an arbitrary target quantile $f\in\RR^4$:
$$
x=\left(\begin{array}{c}4.5 \\1.2 \\10.1 \\8.9\end{array}\right) \,,\quad \Pi_x = \left(\begin{array}{cccc}0 & 1 & 0 & 0 \\1 & 0 & 0 & 0 \\0 & 0 & 0 & 1 \\0 & 0 & 1 & 0\end{array}\right) \,,\quad f=\left(\begin{array}{c}0 \\1 \\3 \\4\end{array}\right) \,, \quad \Phi_f(x) = \Pi_x f =\left(\begin{array}{c}1 \\0 \\4 \\3\end{array}\right) \, .
$$

\section{Supervised quantile normalisation (SUQUAN)}

The QN transform is defined for any arbitrary target quantile $f$ by (\ref{eq:qn}). After QN our $n$ samples $x_1, \ldots, x_n$ therefore become $n$ vectors $\Pi_{x_1} f , \ldots , \Pi_{x_n} f$, amenable for further analysis. We propose that instead of separating the tasks of choosing a "good" target quantile for QN on the one hand, and  analysing the normalised data for some application on the other hand, we couple the two problems and optimise the target quantile in order to better solve the subsequent data analysis problem. 

Let us now instantiate this general principle to the problem of estimating a linear model after QN normalisation; this is useful, for example, when one wants to build a prognostic model for cancer from gene expression data, or classify images based on their content. A linear model with weights $w \in \RR^p$ and offset $b\in\RR$ applied after quantile normalisation with target quantile $f\in\RR^p$ takes the form
\begin{equation}\label{eq:lm}
\forall x\in\RR^p\,,\quad F_{w,b,f}(x) = w^\top \Phi_f(x) + b\,.
\end{equation}
Given samples $x_1,\ldots,x_n$, let us consider a standard procedure where the parameters $(w,b)$ of the linear model are estimated by penalised empirical risk minimisation, i.e., solve an optimisation problem of the form
\begin{equation}\label{eq:erm}
\min_{w,b} \frac{1}{n} \sum_{i=1}^n \ell_i\left(F_{w,b,f}(x_i)\right) + \lambda \Omega(w)\,,
\end{equation}
where $\ell_i$ is a loss function for sample $i$, such as the squared loss $\ell_i(u)=(y_i-u)^2$ for a regression problem with response output $y_i\in\RR$, or the logistic loss $\ell_i(u) = \log\left(1+\exp(-y_i u )\right)$ for a binary classification problem with response output $y_i\in\{-1,1\}$, $\Omega$ is a penalty function such as the $\ell_1$ or $\ell_2$ norm, and $\lambda\geq 0$ is a regularisation parameter. Note that we can rewrite the regularised problem (\ref{eq:erm}) as a constrained optimisation problem:
\begin{equation}\label{eq:erm2}
\min_{(w,b)\in\Wcal\times\RR} \frac{1}{n} \sum_{i=1}^n \ell_i\left(F_{w,b,f}(x_i)\right) \quad\text{where} \quad \Wcal = \left\{ w \in\RR^p\,:\, \Omega(w) \leq C\right\} \,.
\end{equation}

Under mild assumptions, such as the convexity of the $\ell_i$'s and $\Omega$ being a norm, both formulations (\ref{eq:erm}) and (\ref{eq:erm2}) are equivalent in the sense that for all $\lambda>0$ there exist a choice of $C\geq 0$ such that (\ref{eq:erm}) and (\ref{eq:erm2}) have the same solution.

Solving (\ref{eq:erm}) or (\ref{eq:erm2}) is a standard problem in machine learning and statistical estimation, and can be done by a variety of algorithms depending on $n$, $p$, and the specific loss and penalty. Instead of just optimising in $(w,b)$ for a fixed target quantile $f$, chosen independently and often arbitrarily, SUQUAN considers $f$ as a parameter of the full process from the raw data to the final linear models, and optimises $f$ jointly with $(w,b)$. For example, the constrained formulation (\ref{eq:erm2}) becomes:
\begin{equation}\label{eq:suquan}
\min_{(w,b,f)\in\Wcal\times\RR\times\Fcal} \frac{1}{n} \sum_{i=1}^n \ell_i\left(F_{w,b,f}(x_i)\right)\,,
\end{equation}
where $\Fcal\subset\RR^p$ is a set of candidate target quantiles.
Note that the only difference between (\ref{eq:erm2}) and (\ref{eq:suquan}) is the fact that that $f$ is optimised in (\ref{eq:suquan}) and not in (\ref{eq:erm2}); obviously this not only impacts the choice of $f$, but also the solution in $(w,b)$ that is usually different between (\ref{eq:erm2}) and (\ref{eq:suquan}). Note also that since SUQUAN optimises the same objective function as (\ref{eq:erm2}) but over more parameters, the objective function is lower at the optimal solution for SUQUAN than at the optimal solution of (\ref{eq:erm2}); this suggests that SUQUAN has more flexibility to fit the training data, but also more chance of overfitting, and therefore that it may require more regularisation to have good generalisation performance compared to (\ref{eq:erm2}). 

Regarding the set of candidate target quantiles $\Fcal$, one possibility is to simply constrain the Euclidean norm of $f$ to ensure that the regularisation in $w$ has an effect, and consider:
$$
\Fcal_0 = \left\{f \in \RR^p\,:\, \frac{1}{p} \sum_{i=1}^p f_i^2 \leq 1\right\} \,.
$$
 A caveat with $\Fcal_0$ is that the target quantile may not be non-decreasing. We therefore consider a second set of bounded non-decreasing candidate target quantiles:
$$
\Fcal_{\text{BND}}=\Fcal_0 \cap \Ical_0\,, \quad \text{where} \quad  \Ical_0 =  \left\{ f\in\RR^p\,:\, f_1\leq f_2\leq \ldots\leq f_p \right\} \,
$$
denotes the set of non-decreasing vectors.
Further constraints regarding the structure of $f$ may also be encoded in $\Fcal$. For example, if we expect the target quantile to be smooth, we propose to consider the following set of non-decreasing and smooth functions \cite{Sysoev2016smoothed}:
$$
\Fcal_{\text{SPAV}}= \left\{ f \in \Ical_0 \,:\, \sum_{j=1}^{p-1} (f_{j+1} - f_j)^2 \leq 1\right\} \,.
$$
Plugging any of $\Fcal_0$, $\Fcal_{\text{BND}}$ or $\Fcal_{\text{SPAV}}$ into (\ref{eq:suquan}) leads to a SUQUAN formulation with different sets of candidates target quantiles. Note that the presence of the non-penalised intercept $b\in\RR$ in (\ref{eq:lm}) ensures that a solution $f$ to (\ref{eq:suquan}) is defined up to a constant; we can therefore constrain without loss of generality $f$ to be centered ($\sum_{i=1}^p f_i=0$) in $\Fcal_0$ and $\Fcal_{\text{BND}}$, since it corresponds to the constant that minimises the Euclidean norm of $f$, as well as in $\Fcal_{\text{SPAV}}$, since the smoothness constraint is invariant to the addition of a constant.

\section{SUQUAN as a matrix regression problem}

 In order to derive practical algorithms and shed light on the underlying optimisation problems for the different SUQUAN formulations, it is useful to rewrite them as equivalent regression problems. For that purpose, let us now combine the definition of QN (\ref{eq:qn}) and of SUQUAN (\ref{eq:suquan}) together. Plugging (\ref{eq:qn}) into (\ref{eq:lm}) and (\ref{eq:lm}) into (\ref{eq:suquan}), we easily get that the objective function of SUQUAN can be rewritten as:
\begin{equation}\label{eq:trick}
\begin{split}
\frac{1}{n}\sum_{i=1}^n \ell_i\left(F_{w,b,f}(x_i)\right) & = \frac{1}{n} \sum_{i=1}^n \ell_i\left(w^\top \Pi_{x_i} f+b\right) \\
&=  \frac{1}{n} \sum_{i=1}^n \ell_i\left(< w f^\top , \Pi_{x_i}>_F +b\right)  \,,
\end{split}
\end{equation}
where $<A,B>_F = \Tr (A^\top B) = \sum_{i,j=1}^p A_{ij} B_{ij}$ is the standard Frobenius inner product between matrices. This reformulation clarifies that SUQUAN can be interpreted as a particular linear regression model after embedding the inputs space onto the space of $p\times p$ matrices. Indeed, let $\Psi : \RR^p \rightarrow \RR^{p\times p}$ be the mapping defined by 
\begin{equation}\label{eq:psi}
\forall x \in\RR^p\,\quad \Psi(x) = \Pi_x\,,
\end{equation}
then plugging (\ref{eq:trick}) and (\ref{eq:psi}) into (\ref{eq:suquan}) we obtain the following expression for SUQUAN:
\begin{equation}\label{eq:suquan3}
\min_{(M,b)\in\Mcal\times \RR} \; \frac{1}{n} \sum_{i=1}^n \ell_i\left(<M , \Psi(x_i)>_F +b\right) \,,
\end{equation}
where 
$$
\Mcal = \Wcal \otimes \Fcal = \left\{ w f^\top\,:\, w\in\Wcal\,,\,f\in\Fcal\right\}\,.
$$
In other words, SUQUAN can be interpreted as a regression problem after embedding input vectors onto permutation matrices, with a rank-1 constraint on the weight matrix $M$ and additional constraints on its left and right singular vectors corresponding respectively to the linear model $w\in\Wcal$ and the target quantile $f \in \Fcal$, up to a scaling factor.

This intriguing interpretation of target quantile optimisation as constrained matrix regression raises several comments.
\begin{itemize}
\item The mapping $\Psi$ in (\ref{eq:psi}) is the well-known \emph{permutation representation} of the symmetric group $S_p$ \cite{Serres1977Linear, Diaconis1988Group}, where each vector $x\in\RR^p$ is seen as a permutation $\pi_x \in S_p$ defined by the ranking of its entries. In particular, this representation is irreducible when restricted to the set $\Sigma=\left\{f\in\RR^p\,:\,\sum_{i=1}^p f_i=0\right\}$ \cite[exercice 2.6]{Serres1977Linear}, which implies that for any quantile $f\in\Sigma$ (in particular any $f$ that solves (\ref{eq:suquan})), the set of quantile normalised vectors $\left\{\Phi_f(x)\,:\,x\in\RR^p\right\}$ spans the full subspace $\Sigma$.
\item Besides the permutation representation, other embeddings of $S_p$ onto $\RR^{p\times p}$ exist and have been proposed in machine learning. For example, \cite{Jiao2015Kendall} considered mapping $x\in\RR^p$ to a $p\times p$ binary matrix with $(i,j)$-th entry equal to $1$ whenever the $i$-th entry of $x$ is smaller than the $j$-th entry, and showed how Frobenius-norm regularised linear models can be estimated efficiently thanks to the kernel trick because the inner product between two $p\times p$ matrices corresponding to two vector embeddings can be computed in $O(p\ln(p))$ with an efficient implementation of the Kendall $\tau$ statistics. It can be observed that the permutation representation $\Psi$ used by SUQUAN is also trivially amenable to benefit from the kernel trick: to compute the inner product between $\Psi(x)$ and $\Psi(x')$ for two vectors $x$ and $x'$, one just needs to sort the entries of each vector independently, in $O(p \ln(p))$, and count in $O(p)$ how many entries are ranked at the same position. However, the permutation representation is extremely sparse ($p$ non-zero values among $p(p-1)$ zeros) and only controlling the Frobenius norm of $M$ (in order to benefit from the kernel trick) may not be sufficient to fight possible overfitting.
\item $\Mcal$ is not a convex set, and SUQUAN is therefore not a convex optimisation problem. A possible variant of SUQUAN would be to relax the rank constraint and replace it for example by a trace norm constraint, which is known to be a natural convex surrogate for the rank \cite{Srebro2005Rank}.
\end{itemize}

\section{Algorithms}

The SUQUAN formulation (\ref{eq:suquan3}) is a nonconvex optimisation problem since the set of rank-1 matrices $\Mcal$ is not convex. To approximatively solve it, we now propose two strategies. The first one, SUQUAN-SVD, does not really attempt to solve (\ref{eq:suquan3}) but instead to directly find a good target quantile $f\in\Fcal_0$ for binary classification problems. The second one aims to find an approximate solution to (\ref{eq:suquan3}) by performing alternate optimisation in $f$ and $w$, as the problem is biconvex.

\subsection{SUQUAN-SVD}

\begin{wrapfigure}{R}{0.41\linewidth}
\begin{minipage}{0.96\linewidth}
\begin{algorithm}[H]
\caption{SUQUAN-SVD}
\label{alg:suquanSVD}
\begin{algorithmic}[1]
\REQUIRE  $(x_1,y_1) , \ldots , (x_n,y_n) \in \RR^p\times\{-1,1\}$
\ENSURE  $f\in\Fcal_0$ target quantile
\STATE $M_{LDA} \leftarrow 0 \in \RR^{p\times p}$
\STATE $n_{+1} \leftarrow |\{i\,:\,y_i = +1\}|$
\STATE $n_{-1} \leftarrow |\{i\,:\,y_i = -1\}|$
\FOR{$i=1$ to $n$}
\STATE Compute $\Pi_{x_i}$ (by sorting $x_i$)
\STATE $M_{LDA} \leftarrow M_{LDA} + \frac{y_i}{n_{y_i}} \Pi_{x_i}$
\ENDFOR
\STATE $(\sigma,w,f) \leftarrow SVD(M_{LDA},1)$
\end{algorithmic}
\end{algorithm}
\end{minipage}
\end{wrapfigure}

In the case where $\Fcal = \Fcal_0$, i.e., when we do not constrain $f$ to be non-decreasing, and $\Omega(\beta) = ||\beta||^2$, then the set $\Mcal$ of candidate matrices in (\ref{eq:suquan3}) is exactly the set of rank-1 matrices. In that case, (\ref{eq:suquan3}) amounts to finding a rank-1 matrix that approximatively solves a linear regression or classification problem. Let us consider the binary classification setting, when the training set is composed of pairs $(x_i,y_i)_{i=1,\ldots,n}$ with $y_i\in\{-1,+1\}$. In that case, a simple linear classifier (without rank constraint) is the one obtained by linear discriminant analysis with identity covariance: $M_{LDA}=\mu_+ - \mu_-$, where $\mu_+$ and $\mu_-$ are respectively the means of the matrices $\Pi_{x_i}$ for the positive and negative classes. Consequently, a good rank-1 candidate classifier is the closest rank-1 matrix to $M_{LDA}$, namely $u\sigma v^\top$ where $u$ and $v$ are the left and right singular vectors of $M_{LDA}$ associated to the largest singular value $\sigma$. Hence we recover a target quantile function by keeping only the first right singular vector of $M_{LDA}$, which can then be used as target quantile for quantile normalising the training points before running any linear classification method. Algorithm~\ref{alg:suquanSVD} summarises the method. Computing $\Pi_{x_i}$ on line 5 involves an $O(p\ln(p))$ sorting of the entries of $x_i$, and therefore computing $M_{LDA}$, which is a linear combination of $n$ permutation matrices, requires $O(np\ln(p))$ operations. Then computing the right largest singular vector (line 8) of  $M_{LDA}$  typically costs another $O(p^2)$ operations using a naive power iteration method. However, if $n \leq p$, we can exploit the fact that the product of a permutation matrix by a vector is just an O(p) operation (just order the vector according to the permutation), so that the power iteration to compute the first singular vector only takes $O(np)$.  Computing the right largest singular vector therefore has an $O(\min(p^2, np))$ complexity. Hence the complexity of SUQUAN-SVD is $O(np\ln(p))$,  which is the same as the complexity of the quantile normalisation.

\subsection{SUQUAN-BND and SUQUAN-SPAV} 
We now focus on approximate algorithms to solve (\ref{eq:suquan3}) in the case where $\Fcal=\Fcal_{BND}$ or $\Fcal = \Fcal_{SPAV}$. Using the biconvexity of (\ref{eq:suquan3}) in $w$ and $f$, we propose an alternate optimisation scheme in $w$ and $f$. Algorithm \ref{alg:suquanBND} summarises the procedure. Starting from an initial non-decreasing target quantile $f_{init}\in\Ical_0$, it outputs a new target quantile $f$ obtained by minimising once (\ref{eq:suquan3}) in $w$ for $f=f_{init}$ fixed, then minimising in $f$ for $w$ fixed. Each alternative optimisation is particularly simple and efficient. For a given $f$, the optimisation in $(w,b)$ amounts to solving a standard linear model optimisation over the samples $\left(\Pi_{x_1}f , \ldots , \Pi_{x_n}f\right)$. For a given $w$, the optimisation in $f$ differs according to the regularisation type. With $\Fcal_{\text{BND}}$, the optimisation in $f$ is an isotonic optimisation problem (because of the constraints in $\Fcal_{\text{BND}}$ that entries in $f$ should be non-decreasing) involving the samples $\left(\Pi_{x_1}^\top w , \ldots , \Pi_{x_n}^\top w\right)$, which we solve by accelerated proximal gradient optimisation, borrowing the pool adjacent violators algorithm (PAVA, \cite{Barlow1972Statistical}) as proximal operator to project onto the set of monotonically increasing vectors in $O(p)$. With $\Fcal_{\text{SPAV}}$, this is a smoothed isotonic optimisation problem via $\ell_2$ regularisation. Again, we solve this problem by accelerated proximal gradient optimisation but this time borrowing the Smoothed Pool Adjacent Violators (SPAV, \cite{Sysoev2016smoothed}) as proximal operator which costs $O(p^2)$ operations; in this case we solve a penalised version (as opposed to a constrained version) of the problem, inducing a second regularisation parameter $\gamma$. Interestingly, the computation of each matrix-vector products $\Pi_{x_i} f$ and $\Pi_{x_i}^\top w$ before each alternative optimisation is just an $O(p)$ operation, after the sample $x_i$ has been sorted once at the first iteration in $O(p\ln(p))$. Indeed, for a given $x$, if we note $order(x)$ the permutation which rearranges the entries of $x$ in increasing order, and $rank(x)$ the ranks of the entries of $x$, then we simply have $(\Pi_x f)_j = f_{rank(x)_j}$ and $(\Pi_x^\top w)_j = w_{order(x)_j}$, for $j=1,\ldots,p$, which we simply denote as $\Pi_x f = f[rank(x)]$ and $\Pi_x^\top w = w[order(x)]$ in Algorithm \ref{alg:suquanBND}. Note that the procedure can be iterated to produce a sequence of target quantiles although we found in our experiments below that the performance did not change significantly after the first iteration. Note also that, contrary to SUQUAN-SVD, this algorithm requires an initial non-decreasing target quantile function. By default we suggest to use the median of the data quantile functions, which is often the default used in standard QN normalisation.

\begin{center}
\begin{minipage}{0.67\linewidth}
\begin{algorithm}[H]
\caption{SUQUAN-BND and SUQUAN-SPAV}
\label{alg:suquanBND}
\begin{algorithmic}[1]
\REQUIRE  $(x_1,y_1) , \ldots , (x_n,y_n), f_{init} \in \Ical_0$, $\lambda\in\RR$
\ENSURE $f\in\Ical_0$ target quantile
\FOR{$i=1$ to $n$}
\STATE $rank_i , order_i  \leftarrow$ sort$(x_i)$
\ENDFOR
\STATE $w,b \leftarrow \underset{w, b}{\text{argmin}} \frac{1}{n} \sum_{i=1}^n \ell_i\left(w^\top f_{init}[rank_i]+b\right) + \lambda ||w||^2$ \\
 (standard linear model optimisation)
\STATE $f  \leftarrow \underset{f\in\Fcal_{BND}}{\text{argmin}} \frac{1}{n} \sum_{i=1}^n \ell_i\left( f^\top w[order_i]+b\right)$ \\
 (isotonic optimisation problem using PAVA as prox)\\
 OR\\
$f  \leftarrow \underset{f\in\Fcal_{SPAV}}{\text{argmin}} \frac{1}{n} \sum_{i=1}^n \ell_i\left( f^\top w[order_i]+b\right)$ \\
 (smoothed isotonic optimisation problem using SPAV as prox)\\
\end{algorithmic}
\end{algorithm}
\end{minipage}
\end{center}

\section{Experiments}

\subsection{Simulated data}

We first test the ability of SUQUAN to overcome unwanted changes in quantile distributions on simulated datasets. For that purpose we fix $f\in\RR^p$ to be the quantile distribution of the normal distribution, and simulate each sample $x_1,\ldots,x_n\in\RR^p$ by randomly permuting the entries of $f$. We then generate binary labels $y_1,\ldots,y_n\in\{-1,1\}$ using the logistic model $P(Y=1\,|\,X=x) = \frac{1}{1+\text{exp}(-w^\top x)}$, where $w$ is randomly sampled from a standard multivariate normal distribution. We then compare four methods to estimate $w$ from $n$ observations:
\begin{itemize}
\item Ridge logistic regression estimated on the correct data $(x_i,y_i)_{i=1,\ldots,n}$.
\item Ridge logistic regression estimated on the corrupted data $(\Phi_g(x_i),y_i)_{i=1,\ldots,n}$, where $g$ is a corrupted quantile distribution.
\item SUQUAN-BND and SUQUAN-SPAV estimated on the corrupted data $(\Phi_g(x_i),y_i)_{i=1,\ldots,n}$.
\end{itemize}
While the true target $f$ quantile is normal, we test four corrupted target quantiles $g$, derived from the cauchy, exponential, uniform and bimodal gaussian distributions. We assess the performance of the estimation by the area under the curve (AUC) on an independently generated test set of 1000 samples. The hyperparameters controlling the $\ell_2$ penalty on $w$ ($\lambda$) and the smoothness penalty on $f$ ($\gamma$) for SUQUAN-SPAV were chosen thanks to an inner 5 times 3-fold cross-validation. The grid of values tested for $\lambda$ ranges from $10^{-5}$ to $10^5$ in log scale and from  $10^{0}$ to $10^4$ in log scale for $\gamma$.

\begin{figure}[hb]
\centering
\includegraphics[scale=0.55]{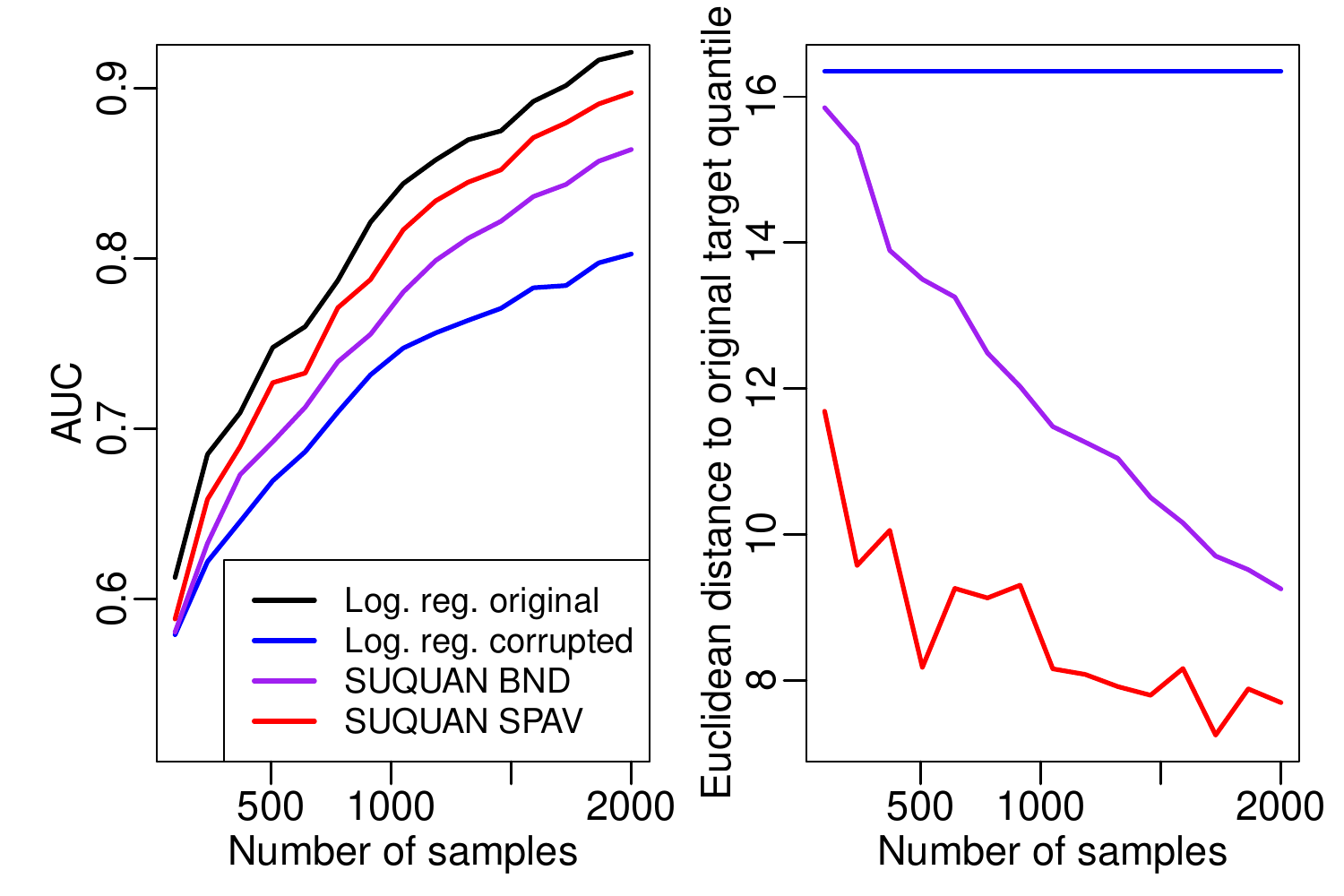}
\caption{Performance on simulated data. The number of features is fixed to p = 1000 while the number of training samples varies from 100 to 2000, and the results are averaged over four experiments with different corrupted quantile functions. The left plot shows the test AUC for logistic regressions applied to the original (black) and corrupted (blue) data as well as SUQUAN-BND (purple) and SUQUAN-SPAV (red). The right plot shows the Euclidean distance between the original target quantile and the target quantile used to corrupt the data (blue), the target quantile learned with SUQUAN BND (purple), and the target quantile learned with SUQUAN-SPAV (red). }
\label{fig:simulations}
\end{figure}

Figure \ref{fig:simulations} shows the performance of the different methods as a function of $n$, the number of training samples. In the case $n \ll p$, all methods perform almost equally badly in terms of AUC, including  linear regressions on the true and on the corrupted datasets. However, SUQUAN-SPAV is able to learn a target quantile which is closer in terms of Euclidean distance to the true target quantile than the initial corrupted target quantile. When the number of samples increases while the number of features is kept fixed, the performances of both SUQUAN-BND and SUQUAN-SPAV clearly outperforms that of linear regression on the corrupted dataset. In particular, the AUC curves show that SUQUAN-SPAV is almost as good as linear regression performed on the true dataset whatever the dimensionality is. Morever, both SUQUAN-BND and SUQUAN-SPAV  improve their estimates of the true target quantile when the number of samples increases. Overall, these results confirm that SUQUAN can improve the performances of a linear model by recovering a good estimate of the true target quantile function, and illustrate the detrimental impact of a bad choice for the target quantile function.

\subsection{CIFAR-10 dataset}

We next test SUQUAN on an image classification task. Since our objective is to study the impact of QN with different target quantile functions, we do not aim to reach state-of-the-art classification results with complex features extracted from images, but instead assess the performance of simple linear models on pixel intensities. Here changing the target quantile can be thought of as a variant of the histrogram matching procedure. We consider the CIFAR-10 benchmark dataset \cite{Krizhevsky2009Learning} which consists of 32$\times$32 tiny color images from 10 different classes. The dataset is divided into 50,000 training images (5,000 of each class) and 10,000 test images (1,000 of each class). To simplify the setting, we consider independently all 45 binary classification problems derived from the 10 classes. For each of these 45 problems, images were first converted to grayscale and represented as vectors of gray intensities. Therefore for each binary problem we have 10,000 training samples, 2,000 test samples, and 1,024 features per image.

\begin{figure}[!ht]
\centering
\begin{subfigure}[t]{0.48\linewidth}
	\centering
	\caption{}\label{CIFAR_boxplot}
	\vspace{-1em}
	\includegraphics[scale=0.56]{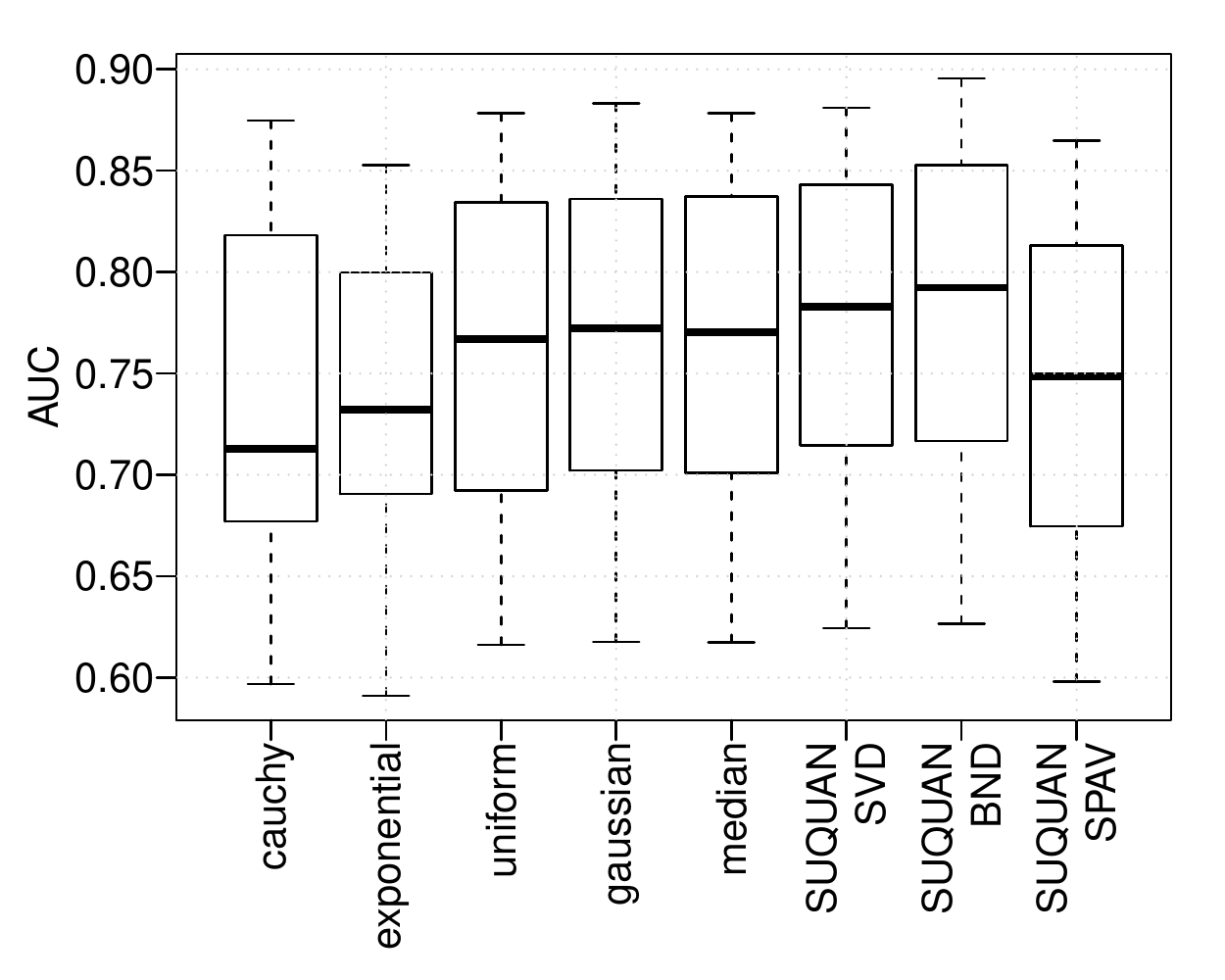}	
\end{subfigure}
\quad
\begin{subfigure}[t]{0.48\linewidth}
	\centering
	\caption{}\label{CIFAR_scatterplot}
	\vspace{-1em}
	\includegraphics[scale=0.69]{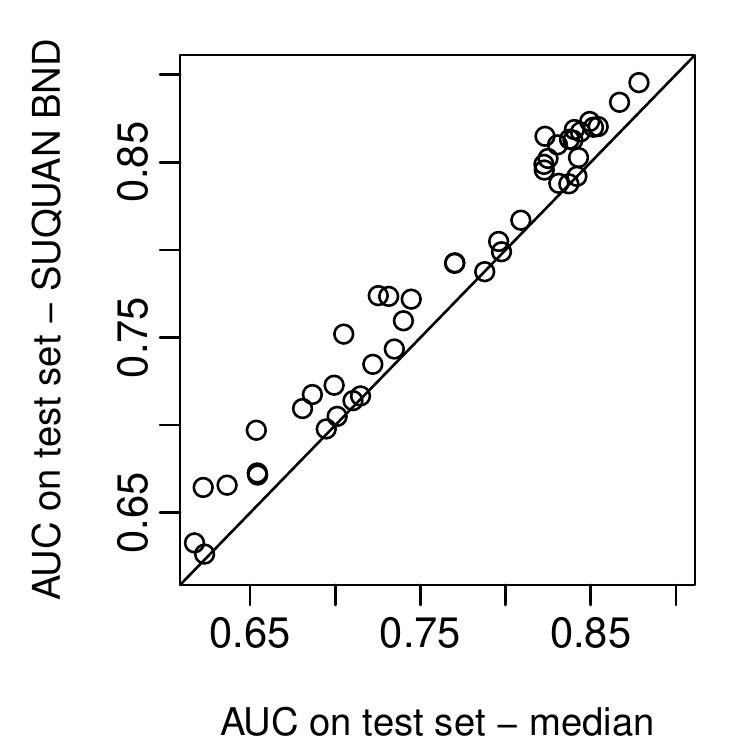}
\end{subfigure}
\caption{ (\subref{CIFAR_boxplot}) Performance on CIFAR-10. Each box-plot summarises the test AUC of a method on the 45 binary classification tasks. For the first seven boxplots on the left, the data was first normalised using a target quantile either drawn from a distribution or estimated by SUQUAN-SVD, and a logistic regression was fitted to the normalised data. The last two cases correspond to directly applying SUQUAN-BND or SUQUAN-SPAV to the data. (\subref{CIFAR_scatterplot}) Comparison of the test AUC obtained with a logistic regression on data previously quantile normalised with the median on the one hand, and SUQUAN-BND on the original data on the other hand. Each point corresponds to one of the 45 binary classification tasks from CIFAR-10.}
\label{CIFAR_boxplot_scatterplot}
\end{figure}

\begin{figure}[]
\center
\includegraphics[scale=0.35]{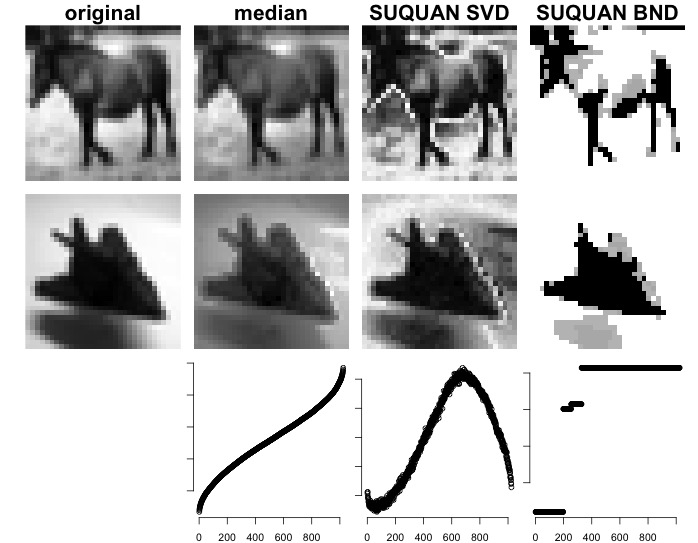}
\caption{Target quantiles for the ``airplane" versus ``horse" binary classification task. The first (resp. second) row represents one sample image from the ``horse" (resp. ``airplane") class in its original form or normalised with the median target quantile across all images , the target quantile from SUQUAN-SVD, or the one learned with SUQUAN BND. The third row shows the shape of the target quantiles $f$ in each case.}
\label{CIFAR_image}
\end{figure}

We compare SUQUAN on these 45 classification tasks to a logistic regression model for which data has been quantile normalised beforehand with various target quantiles. Among these target quantiles we test the median of the empirical distribution of the samples, the target quantile derived from the uniform distribution which amounts to performing histogram matching, as well as the target quantiles derived from the cauchy, exponential and gaussian distribution in order to have diversity in the target quantiles chosen. SUQUAN as well as the logistic regression are fitted with an $\ell_2$ penalty on the weights $w$. Hyperparameters are selected using a 5 times 3 fold cross-validation on the train set. The grid of values tested for $\lambda$ ranges from $10^{-5}$ to $10^5$ in log scale and from  $10^{0}$ to $10^4$ in log scale for $\gamma$.

The distributions of test AUC obtained for each method across all 45 classification problems are shown in Figure~\ref{CIFAR_boxplot}. SUQUAN-BND yields the best average performance and outperforms all logistic regression models learned with fixed target quantiles. Moreover, if we compare the performances of SUQUAN-BND to that of the logistic regression with the median as target quantile (Figure \ref{CIFAR_scatterplot}), we see that the improvements yielded by SUQUAN-BND are consistent across datasets. These observations therefore confirm the benefit  of optimising the target quantile at the same time as the model weights, and support the idea that fixing a pre-defined target quantile can hurt the performance of a linear model. Interestingly, the simplified version of SUQUAN, i.e., SUQUAN-SVD, also creates a target quantile function which outperforms all other fixed target quantiles. In order to illustrate what the learned target quantiles look like for both SUQUAN-BND and SUQUAN-SVD, we show in Figure \ref{CIFAR_image} the normalised images from the `horse' versus `airplane' classification task according to the different methods. We note two things: first, as the target quantile learned with SUQUAN-SVD can be non-monotonic, black pixels in the original image can become white in the normalised image and conversely. Interestingly here this inversion tends to occur at the edges of the objects and therefore plays a role which mimics a simple edge detector; second, SUQUAN-BND learns a target quantile with only few steps, and therefore tends to `binarise' the image, which probably brings out salient features. Finally, we also observe that SUQUAN-SPAV has bad performances on these 45 binary classification tasks, suggesting that the smoothness constraint on the target quantile is detrimental in this case. We hypothesise this may be due to the inherent structure of images, and also to the fact that in a $n \gg p$ setting, constraining the model too much is not necessary.

\subsection{Gene expression data}

Genomic data are often subject to many types of unwanted variations that corrupt the recorded data, including but not restricted to sample preparation protocols, temperature, or measurement tools. To test the relevance of SUQUAN in this context, we focus on the problem of breast cancer prognosis from gene expression data, and collected 4 publicly available datasets describing gene expression profiles in human breast cancer tumors together with survival information from the GEO database \cite{Barrett2011NCBI}. For each of these 4 datasets we retrieved the raw data (CEL files) which we summarised (using the median polish procedure) to obtain gene expressions. Each dataset contains the expression level of 22,283 genes measured using the same microarray technology and the number of breast cancer patients (or samples) ranges from 106 to 271 patients (Table~\ref{tab:GSE}). In each dataset, we split the patients into two classes: those who relapsed within 6 years of diagnosis and those who did not. The precise description of the datasets is summarised in Table~\ref{tab:GSE}. The problem is therefore to predict the class of a patient (relapse or not) given its gene expression values, which is a binary classification task.

\begin{table}[ht]
\caption{Gene expression datasets used in this study (the dataset name corresponds to the accession number in the GEO database)}
\label{tab:GSE}
\begin{center}
\begin{small}
\begin{sc}
\begin{tabular}{c c c c c}
\hline
Dataset name &  \# patients & \# positives & \% positives \\
\hline
GSE1456 & 141 & 37 & 0.26  \\
GSE2034 & 271 & 104 & 0.38 \\
GSE2990 & 106 & 32 & 0.30 \\
GSE4922 & 225 & 73 & 0.32 \\
\hline
\end{tabular}
\end{sc}
\end{small}
\end{center}
\end{table}

We again compare the performances of SUQUAN to that of logistic regression on previously quantile normalised data with various target quantiles namely cauchy, exponential, uniform, gaussian and median. We also fit logistic regressions on the raw data and on the data preprocessed with Robust Multi-Array Average (RMA, \cite{Irizarry2003Exploration}). RMA is a widely used preprocessing method for gene expression microarrays which notably includes a background correction step and a quantile normalisation step with the median as target quantile. Experiments are performed in a 5-times 3-fold external cross-validation setting and the performances reported are the average over these 15 folds. Both models (SUQUAN and logistic regression) are fitted with an $\ell_2$ norm penalty on $w$. Hyperparameters are optimised by 5-times 3-fold inner cross-validation. The grid of values tested for $\lambda$ ranged from $10^{-5}$ to $10^1$ in log scale and from  $10^{0}$ to $10^4$ in log scale for $\gamma$.

\begin{table}
\caption{AUC for SUQUAN and logistic regression with various data normalisation procedures applied to four gene expression datasets.}
\label{tab:perf}
\begin{center}
\begin{small}
\begin{sc}
\resizebox{\linewidth}{!}{%
\begin{tabular}{l|ccccccc|cccr}
\hline
& \multicolumn{7}{c|}{logistic regression} & \multicolumn{3}{|c}{suquan}\\
& raw & rma & cauchy & exp.  & unif. & gaus. & median & svd & bnd & spav\\
\hline
GSE1456 & 65.94 & 68.73 & 59.56 & 68.86 &  68.72 & 69.00 & 69.06 & 57.60 & 71.44 & 69.60\\
GSE2034 & 74.52 & 75.42 & 61.91 & 74.53 &  75.22 & 76.45 & 74.92 & 52.61 & 70.50 & 76.11\\
GSE2990 & 57.01 & 60.43 & 54.72 & 61.25 &  56.25 & 58.66 & 59.72 & 52.51 & 59.22 & 59.94\\
GSE4922 & 58.52 & 58.86 & 55.24 & 58.81 & 55.66 & 60.01 & 59.18 & 52.39 & 61.82 & 61.41\\
\abovestrut{0.20in}
Average & 64.00 & 65.86 & 57.86 & 65.86 & 63.96 & 66.03 & 65.72 & 53.78 & 65.75 & \textbf{66.77}\\
\hline
\end{tabular}
}
\end{sc}
\end{small}
\end{center}
\end{table}

\begin{figure}[ht]
\begin{center}
\includegraphics[scale=0.55]{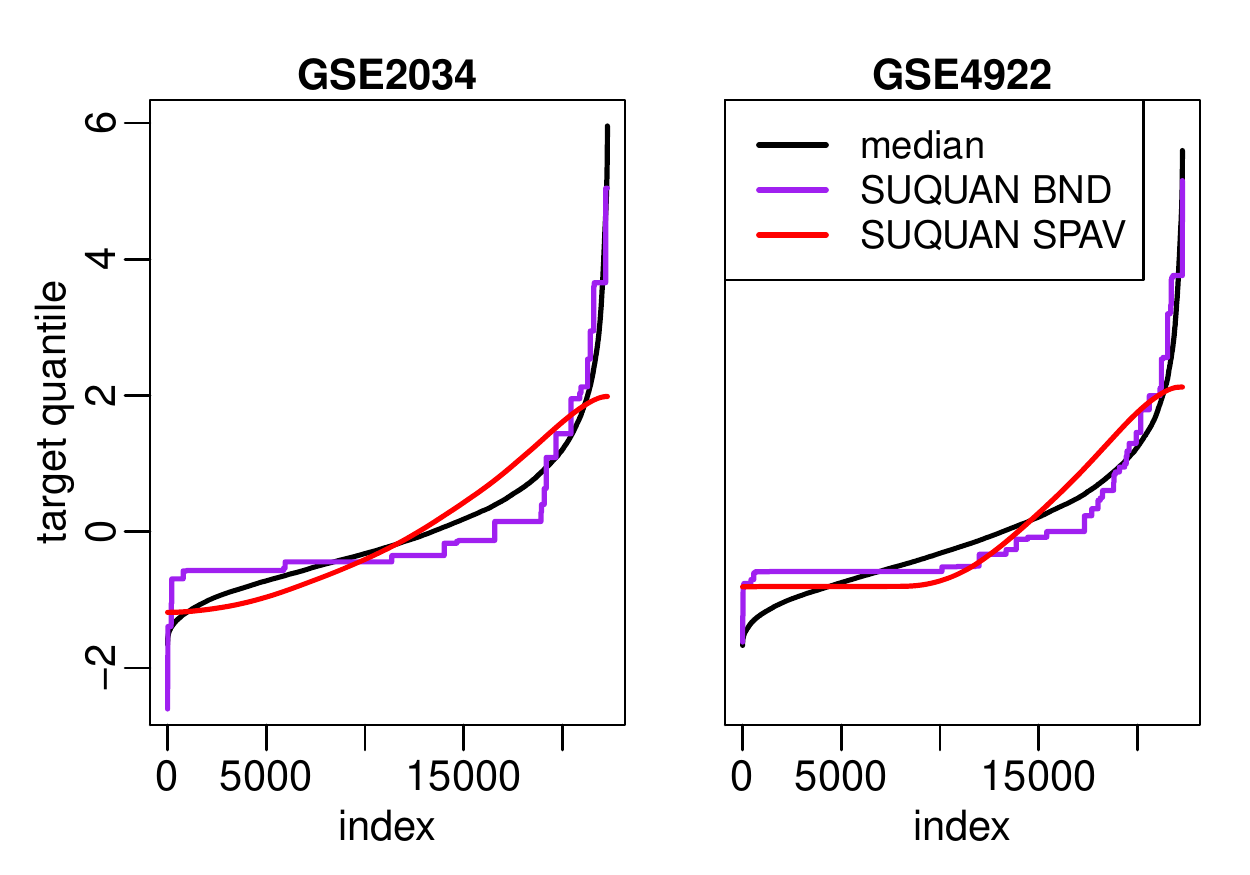}
\caption{Example of target quantiles learned for two gene expression datasets and an arbitrary split in train/test sets.}
\label{fig:GSE}
\end{center}
\end{figure}

Table~\ref{tab:perf} summarises the performance of each method on each dataset. Looking at the mean performance across the four datasets, we observe that the performance of the logistic regression varies according to the target quantile used, which underlines the fact that the choice of the target quantile is important. In particular, RMA (65.86) is one of the top performing preprocessing methods along with quantile normalisation with the median (65.72), exponential (65.86) and gaussian (66.03) target quantiles. Moreover, SUQUAN-SPAV (66.77) outperforms all other methods on average. This increase in performance is significant according to a one-sided paired Wilcoxon signed rank statistical test (P-value $\leq 5 \times 10^{-2}$) for all logistic regressions except those fitted with RMA and the gaussian as target quantile for which the P-values $P=6.9 \times 10^{-2}$ and $P=5.9 \times 10^{-2}$ are just above the significance threshold of $5\%$. We would like to mention that for cancer prognosis from gene expression data, it is very unlikely that any method will ever outperform the baseline by more than a few percents. Illustrations of typical target quantiles learned with SUQUAN-SPAV are shown on Figure \ref{fig:GSE}. Interestingly, while in the large $n$ small $p$ configuration (i.e on CIFAR) SUQUAN-BND was the best method, here in a small $n$ large $p$ configuration SUQUAN-SPAV is better on average than SUQUAN-BND. This may be due to the fact that the smoothness constraint on $f$ which is  implemented in SUQUAN-SPAV is a useful additional regularisation to prevent overfitting. Finally, we observe that SUQUAN-SVD is by far the worst method on these gene expression datasets, probably due to numerical instabilities when computing the singular vectors of large sparse matrices that appear in the $n \ll p$ setting.

\section{Discussion}
QN is an ubiquitous normalization method used throughout several application fields to remove unwanted variations in the recorded data before performing any analysis. However, the choice of the target quantile function is most often empirical and driven by field-specific standard choices. We presented a model, SUQUAN, that allows to learn the optimal target quantile function while performing a given task such as classification or regression. We showed that SUQUAN can be interpreted as a constrained matrix regression problem where sample vectors are embedded as permutation matrices. 

The idea of optimizing the target quantile function jointly with other parameters lends itself well to further investigations. For example, by changing the objective function of SUQUAN, one may consider other applications such as optimizing the quantile function in order to improve clustering or visualization of the data after QN, or the signal-to-noise ratio to detect differentially expressed genes. Regarding SUQUAN itself, a better understanding of the statistical properties of learning low rank linear models on the permutation representation of the symmetric group, as well as extensions from rank-1 to low rank matrices, are interesting future work.

Another remaining challenge is to develop non linear extensions of SUQUAN, using for example kernels. Such an extension is not straightforward since the optimization is not regularized by an L2 norm of the linear model, which would be needed for a simple ` kernel trick ' extension. Instead it is regularized by a rank constraint on the model, which we empirically observed to be crucial. Another way to think about ` kernelizing' the model would be to replace the representation of a permutation as the permutation matrix by something else (i.e., keep the model linear but in another representation), which leads to the question of defining more general kernel or feature representation for the symmetric group, a topic of broader interest in machine learning.

\bibliographystyle{plain}
\bibliography{icml_2017}

\end{document}